# Evaluating BERT-based Pre-training Language Models for Detecting Misinformation

Rini Anggrainingsih[a,b], Ghulam Mubashar Hassan[b] and Amitava Datta[b]

[a]*Informatics Dept, Faculty of Mathematics and Natural Sciences, Sebelas Maret University, Indonesia*
[b]*Department of Computer Science and Software Engineering, University of Western Australia*

## ABSTRACT

The rapid growth of internet technology and social media platforms has led to rapid changes in communication by providing many conveniences to users. Despite the comforts provided, it is challenging to control the quality of information due to the lack of supervision over all the information posted on online media. For instance, there are many problems in society because of the numerous rumours and the quick spread of fake news through online media. Manual checking is almost impossible given the vast number of posts made on online media and how quickly they spread. Therefore, there is a need for automated rumour detection technique to limit the adverse effects of spreading misinformation. Previous studies mainly focused on finding and extracting the significant features of text data and used standard supervised learning techniques or deep learning approaches to identify misinformation. However, extracting features is time-consuming and not a highly effective process due to the nonavailability of some features, affecting the effectiveness and performance of the detection models. Therefore, this study proposes the BERT-based pre-trained language models to encode text data into vectors and utilise neural network models to classify these vectors to detect misinformation. Furthermore, the performance of different language models (LM) containing different number of trainable parameters were compared, including RoBERTa, BERT, and DistillBERT. The proposed technique is tested on different twitter and fake news datasets to represent short and long text data respectively. The study also proposes a standardised distribution ratio for taining, validation and testing of the model. The result of the proposed technique has been compared with the state-of-the-art techniques on the same datasets. The results show that the proposed technique performs better than the state-of-the-art techniques. Moreover, the choice of LM either with significantly large or small number of parameters does not affect the performance considerably. The experiments also shows that a simple classifier performs better than a complex neural network which shows the vectors produced by LMs have enough information to classify these vectors to detect misinformation. We also tested the proposed technique by combining the datasets. The results demonstrated the large training and testing size of data considerably improves the performance of the technique. Therefore, it is suggested that the dataset, splitting data, and classification techniques must be considered carefully to analyse performance of the solutions.

## 1. Introduction

The rapid growth of internet technology and online social media has revolutionised communication. People can freely and easily access the latest news, share, and express their opinions on social media. However, these conveniences may promote the creation and spread of false information, such as rumours and fake news that cause social problems, including financial loss, public panic, personal or community reputation defacement, and human life threats. Recently, spreading rumours and fake news on online media platforms has become a serious issue. It is nearly impossible for ordinary people to distinguish rumour and non-rumour information from most online platforms. Therefore, there is a need for automated detection technology to identify rumours or fake news on online social media.

Castillo et al. (Castillo et al., 2011), who were known as pioneers in detecting credible information research area, used feature engineering and machine learning techniques to detect rumours on Twitter and provided a baseline result for other studies. Other studies then attempted to improve the accuracy of detecting misinformation by discovering the essential features from the context and content of text data. Furthermore, some other studies compared the performance of various machine-learning algorithms for detecting rumours, such as Random Forest, Support Vector Machine, K-Nearest Neighbour, Naïve Bayes Classifier, and Logistic Regression (Ito et al., 2015; Zubiaga et al., 2016; Hassan and Haggag, 2018; Herzallah et al., 2018). Similarly, feature engineering and machine learning techniques are also widely used to detect fake news on online media (Ahmed et al., 2018; Dong et al., 2019a; Reis et al., 2019; Al-Ahmad et al., 2021).

---

* This document is the results of the research project funded by Sebelas Maret University, Indonesia.

✉ rini.anggrainingsih@staff.uns.ac.id, rini.anggrainingsih@research.uwa.edu.au (R. Anggrainingsih); ghulam.hassan@uwa.edu.au (.G.M. Hassan); amitava.datta@uwa.edu.au (.A. Datta)
ORCID(s): 0000-0002-8231-4446 (R. Anggrainingsih)

Following the success of the neural network based approaches in various classification problems, most recent studies leveraged different neural network based techniques for detecting rumours and fake news on online media, including Recurrent Neural Network (RNN) (Ma et al., 2015; Ruchansky, 2017; Alkhodair et al., 2020), Convolutional Neural Network (CNN) (Yu et al., 2017; Xu et al., 2020; Santhoshkumar and Babu, 2020; Bharti and Jindal, 2021; Wang, 2017), and Long Short Term Memory (LSTM) network (Ajao, 2018; Kochkina et al., 2018; Karimi et al., 2018). Some studies used different methodologies to detect rumours on Twitter, such as reinforcement learning (Zhou and Li, 2019) and graph neural networks (Yuan et al., 2019; Wu et al., 2020b; Lu and Li, 2020; Wang et al., 2020).

The emergence of Bidirectional Encoder Representation from Transformer (BERT) as a pre-trained language model (LM) in 2018 by Devlin et al. (Devlin et al., 2018) has significantly improved the area of Natural Language Processing (NLP). BERT can understand a word's contextual meaning by considering words from the left and the right sides. Furthermore, it can be finetuned to deal with other NLP-related problems. Pre-trained LMs have been applied on various types of NLP tasks, such as text summarisation (Moradi et al., 2020; Ma et al., 2021; Liu and Lapata, 2019), text generation (Zhang et al., 2019; Chen et al., 2019; Qu et al., 2020), and text classification (Li et al., 2019; Sun et al., 2019; González-Carvajal and Garrido-Merchán, 2020; Rietzler et al., 2019; Yu and Jiang, 2019). BERT has become the current research trend and marks a new era of NLP because of its high-performance results. Despite the benefits and ability of BERT, some issues are attributed to its complexity. Thereofore, some studies attempted to improve the LM's performance based on BERT's architecture and introduced other LM, such as RoBERTa (Liu et al., 2019) and DistilBERT (Sanh et al., 2019) which were computationally less complex.

Motivated by the ability of pre-trained LM to classify text, this study aims to examine and compare the output performance of three different BERT-based language models when combined with neural network classifiers to detect misinformation. Different LMs used are RoBERTa, BERT and DistilBERT. Each LM is finetuned on labelled datasets and encoded into vectors. Afterward, these vectors were used to train the neural network-based techniques to detect misinformation. The model was validated on short and long text datasets (Tweets and fake news) which are: Pheme1 and Pheme2 datasets from (Zubiaga et al., 2016; Kochkina et al., 2018), Twitter15 and Twitter16 datasets (Ma et al., 2015) and COVID-19 Fake News dataset from Mendeley (Koirala, 2021). The contributions of this study are:

- Proposing a new model to improve rumour detection performance based on combining finetuned LM and neural network.

- Thoroughly evaluating the performance of three different variants of the proposed model in detecting rumours and fake news on four different datasets.

- Introducing a new benchmark testing procedure where all datasets are distributed in the same manner for training, validation and testing.

- Demonstrating that the proposed model outperforms the recent state-of-the-art techniques in rumour and fake news detection.

The rest of the paper is structured as follows: Sections 2 and 3 review the related works on rumour detection and the details of the proposed model on utilising pre-trained LM and neural networks to detect rumour and fake news, respectively. The experiments are discussed to investigate the model's performance in Section 4. Finally, Section 5 summarises the study.

## 2. Related Work

This section explains spreading rumours and fake news on online social media. Later, recent studies related to both rumour and fake news detection techniques are categorised and discussed systematically in different subsections.

## 2.1. Rumour and Fake News Detection

Though studies give different definitions, rumor and fake news are often used interchangeably. A rumour is a story that seems credible but complicated to verify since some part of the information could remain unverified (Bondielli and Marcelloni, 2019). In contrast, fake news is false information that spreads and appears credible to gain biased public opinion on particular issues (Meel and Vishwakarma, 2020). The lack of supervision over social media posts makes rumours and fake news propagate quickly in the society.

Spreading rumours and fake news has become a serious concern due to the fast growth of internet technology and online media platforms. These issues often lead to anxiety or scepticism and attract readers to share them without verification (Zubiaga et al., 2016; Pamungkas et al., 2019). Most rumour and fake news on different topics spread through online media daily and influence people's opinions. Therefore, there is an urgent need to detect rumours or fake news automatically since manual fact-checking is a challenging and time-consuming process.

## 2.2. Traditional Machine Learning-based Approaches

Most of the earlier studies regarding information credibility focused on finding and extracting the significant features from the sources, such as the text contents and user details (Castillo et al., 2011; Ito et al., 2015; Herzallah et al., 2018; Hassan and Haggag, 2018; Ghenai and Mejova, 2017; Chatterjee et al., 2018; Ahmed et al., 2018; Dong et al., 2019a; Reis et al., 2019; Al-Ahmad et al., 2021). Standard machine learning techniques, such as Random Forest, Conditional Random Field, Support Vector Machine, Naïve Bayes, and K-Nearest Neighbour were used to detect the authenticity of the tweets.

Similarly, researchers handled fake news detection problems. Recently, Al-Ahmad et al. conducted a comprehensive study. The study applied and compared selected evolutionary classification models, including Particle Swarm Optimisation (PSO), Genetic Algorithm (GA), and Salp Swarm Algorithm (SSA), to detect fake news related to COVID-19 issues (Al-Ahmad et al., 2021). This study reported that the best results were obtained using K-Nearest Neighbor and Genetic Algorithm techniques.

Overall, feature extraction is a fundamental step of a machine learning approach. However, manually extracting features is a time-consuming process that becomes more challenging as some features are missing due to a user's security settings. This condition affects the effectiveness of the feature-based approach.

## 2.3. Deep Learning-based Approaches

Deep learning with neural network approaches has shown promising results in various text classification problems. In the context of false information detection, the widely implemented neural network framework approaches on detecting rumour are Recurrent Neural Network (RNN) (Ma et al., 2015; Ruchansky, 2017; Alkhodair et al., 2020), Long-Short Term Memory (LSTM) (Ajao, 2018; Kochkina et al., 2018; Karimi et al., 2018) and Convolutional Neural Network (CNN) (Yu et al., 2017; Xu et al., 2020; Santhoshkumar and Babu, 2020; Bharti and Jindal, 2021; Wang, 2017). Recent studies that used neural network-based hybrid approaches have achieved the new state-of-the-art results, which are considered in this study as benchmark results and compared to the performance of the proposed methods.

Wu and Rao worked on rumour detection by utilising interaction between the features of rumours. The study proposed models called "Adaptive Interaction Fusion Networks (AIFN)" and "Gated Adaptive Interaction Networks (GAIN)" to identify true and false rumours on Twitter (Wu and Rao, 2020). These techniques were used to identify interaction features among the tweets and capture semantic conflict in posts and comments. The study evaluated their models on Pheme2 dataset (Kochkina et al., 2018). Later a "Decision Tree-based CoAttention model (DTCA)" was proposed to extend this study by utilising the interaction of credible comments as evidence to detect the truth of the tweet (Wu et al., 2020a). The study reported 82.46% accuracy and 82.5% F1-score.

Wang et al. (Wang et al., 2020) aimed to detect rumour by considering a background hidden knowledge in the post's text content and proposed a "Knowledge-driven Multimodal Graph Convolutional Network". This approach combined textual, conceptual and visual data to represent the semantic information into a unified framework for fake news identification. It used Pheme1 dataset (Zubiaga et al., 2016) to validate the model. The study reported above 87% for all evaluation parameters including, accuracy, precision, recall and F1-score.

Lu and Li (Lu and Li, 2020) aimed to predict whether the source of tweets was fake by using user profiles and social interactions. A Graph-Aware CoAttention Network (GCAN) was proposed based on neural network models to depict the interaction among users and capture the relationship between the source tweet, its propagation and user interaction to generate a prediction. The model was evaluated using Twitter15 and Twitter16 datasets (Ma et al., 2015). Accuracy of 87% and 90% on Twitter15 and Twitter16 datasets were reported, respectively.

Wang and Guo encoded tweets' sentiment information and word representation with a two-layer Cascaded Gated Recurrent Unit (CGRU) to detect rumours (Wang and Guo, 2020). The model was validated using Twitter16 datasets (Ma et al., 2015). The model achieved 88.5% accuracy that outperformed earlier studies. Another study (Santhoshkumar and Babu, 2020) classified tweets as rumour or non-rumour using a Dual Convolutional Neural Network (DCNN) technique using the inherent features of the information set. Their study used Twitter15 and Twitter16 datasets (Ma et al., 2015) to evaluate the model and reported F1 scores of 86% and 87%, respectively.

## 2.4. BERT-based Language Model

The emergence of Language Models (LM) and transformers significantly improved the solutions related to Natural Language Processing (NLP). BERT is a pre-trained language model trained on vast unlabelled data from the BooksCorpus that includes 800 million words and English Wikipedia with 2,500 million words without any genuine training objective. Therefore, it can be finetuned for many NLP tasks. BERT can understand a word's contextual meaning by considering words from both the left and the right sides. Furthermore, it can represent words and sentences that are converted into numeric vectors (Devlin et al., 2018). There are two standard architectures of BERT: $BERT_{BASE}$ and $BERT_{LARGE}$. In general, $BERT_{BASE}$ has 12 encoder layers with 110 million parameters and 768 hidden layers, and $BERT_{LARGE}$ has 16 encoder layers with 340 million parameters and 1,024 hidden layers.

Despite the benefit and ability of the pre-trained language model, there are issues regarding training, memory consumption, and computational power due to a large amount of training data. Therefore, some researchers introduced different LMs to address these issues, such as RoBERTa ((Liu et al., 2019)) to address a training model issue and DistilBERT ((Sanh et al., 2019)) to deal with the memory and computational inefficiency problems.

Liu et al. examined the effect of hyperparameter tuning and training size on BERT architecture (Liu et al., 2019). The results showed that BERT was significantly under-trained. Therefore, an improved BERT training model called Robustly Optimised BERT Approach (RoBERTa) was proposed. The training procedures were modified, including; 1) training the model with a larger dataset; 2) removing the next sentence prediction objective; 3) training in a more extended sequence; and 4) changing the masking pattern dynamically to the training data. RoBERTa was trained by following $BERT_{LARGE}$ architecture. However, unlike BERT, which initially trained using 16GB of the dataset, a more extensive training dataset containing 160GB of text was used to train RoBERTa. In addition, RoBERTa was trained for more iterations of 300,000, which was later extended to 500,000. RoBERTa consistently outperformed BERT in all individual tasks on the standard GLUE benchmark (Wang et al., 2018).

Although BERT and RoBERTa, are leading in the current NLP research, the large models raise some issues regarding the computational cost and requirements. Sanh et al. proposed a lighter and faster LM based on BERT architecture using knowledge distillation to compress the model to deal with computational issues (Sanh et al., 2019). Furthermore, they trained the compressed model to reproduce the behaviour of the large model. The compressed model is called DistilBERT. It has 66 million parameters, 40% fewer than BERT and trains 60% faster than BERT. The results showed that DistilBERT models could achieve good results on several NLP tasks and are computationally light enough to run on mobile devices.

Some researchers in the information credibility area who applied BERT-based LMs in classifying misinformation have reported that they obtained an excellent performance by using LMs instead of using standard machine learning methods (Anggrainingsih et al., 2021; Kaliyar et al., 2021; Gupta et al., 2021). Furthermore, other researchers compared LMs' performance to evaluate the cross-source failure problem in the current misinformation detection methods. They focused on finding generalisable representation so that the classification model would be more applicable in real-world data (Huang et al., 2020). They examined the cross-source generalisability by choosing one dataset as a training set and treating other datasets as testing sets.

Pelrine et al. evaluated the performance of a few pre-trained Language Models on detecting rumours (Pelrine et al., 2021). However, they took a different direction to facilitate a solid standard benchmark by using different distribution of data on each dataset just same with the previous studies to make a fair benchmark comparison. The LMs were finetuned, and a single layer perceptron was used as a classifier in their study. The results showed that the performance of the proposed method was better than the state-of-the-art models. Therefore, this study is included as a current state-of-the-art model and compared with the proposed methods.

# 3. Material and Method

This section describes the proposed models and datasets used to validate the proposed models and explains the experimental steps used in this study. This study focuses on comparing and evaluating the performance of three different size BERT-based LMs, including RoBERTa, BERT$_{BASE}$, and DistilBERT on distinguishing rumour and non-rumour tweets, classifying true and false-rumours, and detecting true and fake-news by using neural network-based classifier. Various datasets were then used to validate the performance of the proposed methods. The proposed method and datasets are explained below.

## 3.1. The Proposed Model

This study utilises the finetuned model of LMs as an encoder to represent text into vectors and applying neural network models as a classifier. The architecture of the proposed method is presented in Figure 1.

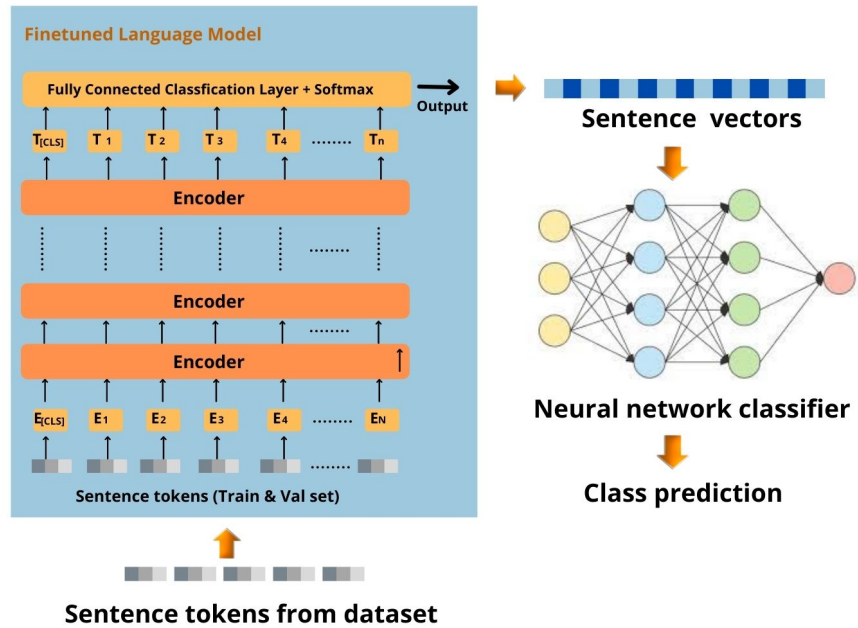

**Figure 1:** The proposed architecture for detecting credibility of information

The finetuned LM was leveraged to encode the sentences into vectors. After the data was encoded into vectors by finetuned LMs, two different neural network techniques were used for classification, including Multilayer Perceptron (MLP) and Residual Network on Convolutional Neural Network (Resnet-CNN). MLP is a neural network that comprises more than one perceptron. Generally, it consists of an input layer to receive the signal, an output for predicting the input, and an arbitrary number of hidden layers between the input and output (Taud and Mas, 2018). This study uses four layers of MLP (4L-MLP) and performed regularisation and dropout on the 4L-MLP model (4L-MLP+Reg+Drop). While ResNet-CNN is an improved model for addressing the vanishing gradient problem that appears on a Convolutional Neural Network (CNN) with deeper layers. A CNN is similar to MLP, though it comprises several layers in its hidden layer, including convolutional, pooling, fully connected, and regularisation layers. These additional layers play an essential role in the success of most deep neural networks. However, a study (He et al., 2016) empirically showed a maximum threshold for the depth of the additional layers of the CNN model. A new neural layer called the residual network was introduced to deal with the problem of deeper networks. ResNet comprises residual blocks, which skip connections to some layers and join the outputs from prior layers to the output of stacked layers. This skipped connections could solve the vanishing gradient problem in CNN and improve neural networks' performance with more layers (He et al., 2016). This study uses 10-layers and 18-layers ResNet-CNN for classification purposes.

## 3.2. Datasets

Both long-text (fake news) and short-text (tweets) datasets were used to validate the proposed method's performance. Regarding the splitting data technique, this study suggests to take the same distribution of data for all datasets to make a standard testing procedure. It is different from many existing studies, that took a different distribution based on the state-of-the-art results of each dataset. The datasets are explained in detail in the following subsections.

### 3.2.1. COVID-19 Fake News Dataset

COVID-19 Fake News Dataset from Mendeley (Koirala, 2021) was selected to represent a long-text dataset. This dataset contains news about COVID-19 issues from December 2019 to July 2020. Each item is labelled as true-news (2,061) and as false-news (1,058). It is called as Covid T/F dataset.

### 3.2.2. Pheme Datasets

There are two types of Pheme datasets: Pheme1 and Pheme2. Pheme1 contains 5,791 tweets about five breaking news topics where 1,969 tweets are labelled as rumours and 3,822 are labelled as non-rumours by journalists (Zubiaga et al., 2016). Pheme2 is the extended version of Pheme1, with four additional news topics, meaning that Pheme2 contained 6,425 tweets about nine news events in total. The journalists verified whether these tweets are rumour (2,402 tweets) or non-rumour(4,023 tweets). Then, rumour tweets are further classified and labelled as true-rumour (1,067 tweets), false-rumour (638 tweets) or unverified-rumour (697 tweets) (Kochkina et al., 2018). However, this study does not consider the data with unverified-rumour label as it is beyond the scope of this study.

Pheme datasets are used in different manner in different state-of-the-art techniques. Therefore we evaluated the proposed models in the same manner to make a fair comparison with existing studies. These detailed sub-datasets are explained below:

- Pheme1 R/NR: It contains 5,791 tweets where 1,969 are labelled as rumour tweets and 3,822 are labelled as non-rumour tweets from Pheme1 dataset.

- Pheme2 T/F: It contains only 1,705 tweets that are marked as true or false rumours and taken from Pheme2 dataset. It consists of 1,067 and 638 tweets labelled as true-rumours and false-rumours, respectively.

- Pheme2 R/NR: It consists of 6,425 tweets from Pheme2 dataset where 2,402 tweets are categorised as rumour tweets while 4,023 are categorised as non-rumour tweets.

### 3.2.3. Twitter15 and Twitter16

Twitter 15 and 16 are publicly available datasets containing 1,490 and 818 tweets. The journalists labelled each tweet as rumour or non-rumour. Afterwards, each rumour tweet is classified and labelled as a true-rumour, false-rumour or unverified rumour (Ma et al., 2015). Similar to the Pheme2 dataset, this study did not consider the data with unverified-rumour labels. This study uses four versions of this datasets, as follows:

- Twitter15 R/NR: It contains 1,490 tweets from the Twitter15 dataset where 1,118 are labelled as a rumour while 372 are labelled as non-rumour.

- Twitter16 R/NR: It consists of 818 tweets from the Twitter16 dataset marked as a rumour (613 tweets) or non-rumour (205 tweets).

- Twitter15 T/F: It contains tweets labelled as true-rumour (375 tweets) or false-rumour (370 tweets) from the Twitter15 dataset.

- Twitter16 T/F: It comprises 410 tweets from Twitter16 dataset, which are marked as true-rumour ( 205 tweets) or false-rumour (205 tweets).

### 3.2.4. Combined Dataset

In order to generalise the performance of the proposed method, we combined the dataset as suggested by (Pelrine et al., 2021). In this study, we approached the problem differently than (Huang et al., 2020) where different datasets are used for training and testing. This study generalised combined short-text datasets with the same label from Pheme2, Twitter15, and Twitter16 datasets to obtain a generalised performance result in classifying rumour and non-rumour tweets and, distinguishing true and false rumour tweets.

**Table 1**
Distribution of labels in the datasets used in this study

| Datasets and labels | COVID-19 Fake News (Koirala, 2021) | Pheme1 (Zubiaga et al., 2016) | Pheme2 (Kochkina et al., 2018) | Twitter15 (Ma et al., 2015) | Twitter16 (Ma et al., 2015) | Combined R/NR | Combined T/F |
|---|---|---|---|---|---|---|---|
| False-news | 1,058 | - | - | - | - | - | - |
| True-news | 2,061 | - | - | - | - | - | - |
| Non-rumours | - | 3,822 | 4,023 | 372 | 205 | 4,600 | - |
| Rumour | - | 1,969 | 2,402 | 1,118 | 613 | 4,133 | - |
| False-rumours | | | 638 | 370 | 205 | - | 1,213 |
| True-rumours | | | 1,067 | 374 | 205 | - | 1,646 |
| *Unverified (not used in this study)* | | | *697* | *374* | *203* | *-* | *-* |

A single combined dataset with rumour and non-rumour label named Combined R/NR was created from Pheme2 R/NR, Twitter15 R/NR, and Twitter16 R/NR to validate the model's performance in classifying rumour and non-rumour tweets. Hence, Combined R/NR dataset comprises 4,600 non-rumour tweets and 4,133 rumour tweets. Similarly, Pheme2 T/F, Twitter15 T/F, and Twitter16 T/F datasets were incorporated into a single dataset named Combined T/F dataset. The dataset consists of 1,213 false-rumour tweets and 1,646 true-rumour tweets. Table 1 describes the more detailed distribution of the datasets used in this study.

### 3.3. Experimental Procedure

The experiments used a RTX 2080 GPU to train each model. The experimental setup for identifying rumours/non-rumours tweets, and true/fake-news using LMs and neural network models is shown in Figure 2.

As a first step, dataset was split into a training, validation, and testing sets. From each dataset, 10% of the data is reserved first for testing. The remaining data was split into 75% and 25% for training and validation sets, respectively. For the combined datasets, 10% of the data was taken from each dataset for the testing-set before blending them into a single Combined T/F dataset or Combined R/NR dataset for training and validation. The rest of combined dataset was split into 3:1 proportions for the training and validation sets, respectively.

The next step was the finetuning of the LM using Huggingface library (Wolf et al., 2019) and the labelled data as an input. Adam optimiser was used while the learning rate was set at 5e-5 with a batch size of eight and experiments were run for ten epochs. The mean pooling layer was set as a pooler to encode the datasets into vectors. Primarily, these vectors were used to train the classifier model.

In the classification step, the learning rate was set at 2e-4, batch size at 512, and the number of maximum epochs selected was 1,000. The Models were trained using Adam optimiser, and cross-entropy as the loss functions. At the end, the models were evaluated by calculating their accuracy, precision, recall, and F1-score using equations(1) to (4):

$$Accuracy(A) = \frac{TP + TN}{TP + TN + FP + FN} \tag{1}$$

$$Precision(P) = \frac{TP}{TP + FP} \tag{2}$$

$$Recall(R) = \frac{TP}{TP + FN} \tag{3}$$

$$F1 = \frac{2(P)(R)}{P + R} \tag{4}$$

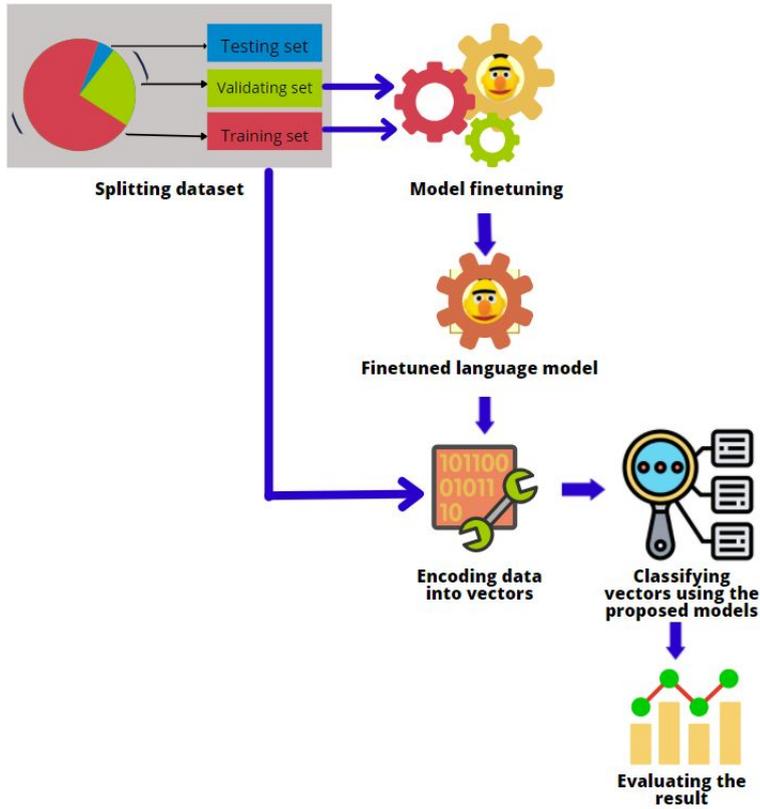

**Figure 2:** The experiment setup to identify rumours/non-rumour tweets and true/fake news using the proposed model.

where TP represents True-positive cases, FN represents False-negative cases, FP represents False-positive cases, and TN represents True-negative cases.

## 4. Results and discussion

This section presents the experimental results of the proposed models and compares them with the results of state-of-the-art models for each dataset. The works which are considered as the state-of-the-art models on each dataset and their performance are presented in Table 2. The comparative performance of all compared models are presented in Table 3 to Table 10. Each table's first and second rows provide details of the state-of-the-art models and their splitting data strategy that achieved the state-of-the-art performance. The best result from each column is in bold, and the second-best and the third-best results are represented in underlined and italic font, respectively.

### 4.1. Comparison of Results with State-of-the-art Methods

Table 3, Table 4, and Table 5 present the experimental results of the proposed methods on classifying rumour or non-rumour on different datasets. Table 3 presents the experimental results and their comparison in distinguishing rumour and non-rumour tweets on the Pheme1 R/NR dataset. It can be observed that all the proposed models using finetuned RoBERTa LM outperformed the state-of-the-art technique of (Wang et al., 2020). The best scores achieved are 88.9% accuracy, 87.9% precision, 87.7% recall and 87.8% F1-score. Although the proposed models attained slightly lower F1-score than (Pelrine et al., 2021) while other metrics were not available for (Pelrine et al., 2021).

Table 4 illustrates the experimental results of the proposed models on Pheme2 R/NR dataset that distinguishes rumour and non-rumour tweets. The observations are similar to Pheme1 R/NR results, where RoBERTa performed

**Table 2**
State-of-the-art results for rumour and fake news detection for each dataset considered in this study.

| Datasets | Studies | Splitting data strategy | Best performance | | | |
|---|---|---|---|---|---|---|
| | | | A (%) | P (%) | R (%) | F1 (%) |
| Pheme1 R/NR | (Wang et al., 2020) | 70:10:20 for train:val:test | 87.56 | 87.62 | 87.65 | 87.64 |
| | (Pelrine et al., 2021) | 70:10:20 for train:val:test | - | - | - | 89.4± 0.3 |
| Pheme2 R/NR | (Dong et al., 2019a) | Not provided | 83.36 | 83.59 | 99.64 | 90.91 |
| Pheme2 T/F | (Wu et al., 2020a) | 70:10:20 for train:val:test | 82.46 | 79.08 | 86.24 | 82.5 |
| | (Pelrine et al., 2021) | 70:10:20 for train:val:test | - | - | - | 93.2±0.9 |
| Twitter15 T/F | (Lu and Li, 2020) | 70:30 for train:test | 87.67 | 82.57 | 82.95 | 82.50 |
| | (Pelrine et al., 2021) | 70:10:20 for train:val:test | - | - | - | 94.4±0.8 |
| Twitter16 T/F | (Lu and Li, 2020) | 70:30 for train:test | 90.84 | 75.94 | 76.32 | 75.93 |
| | (Pelrine et al., 2021) | 70:10:20 for train:val:test | - | - | - | 95.7±2.8 |
| Twitter15 R/NR | (Ma et al., 2019) | Not provided | 86.3 | - | - | - |
| | (Santhoshkumar and Babu, 2020) | 60:40 for train:test | 86 | - | - | - |
| Twitter16 R/NR | (Wang and Guo, 2020) | 75:25 for train:test | 88.5 | - | - | - |
| | (Santhoshkumar and Babu, 2020) | 60:40 for train:test | 87 | - | - | - |
| Covid T/F | (Al-Ahmad et al., 2021) | Not provided | 75.43 | 66.22 | 56.33 | 60.9 |

Abbreviations: A:Accuracy, P: Precision, R: Recall, F1: F1 score

generally better than other LMs. This can be contributed to the fact that RoBERTa has higher number of parameters than other LMs. It can also be observed that 4L-MLP with regularisation and dropout, outperformed both ResNet-CNN classifiers. However, the difference in classifier's performance between RoBERTa, BERT, and DistillBERT on the Pheme2 R/NR dataset is small, and it only varies from 1% to 3%. In summary, all the proposed models outperform the state-of-the-art model in accuracy and precision, while are similar on F1 score.

**Table 3**
Comparison of the proposed method with state-of-the-art techniques for Pheme1 R/NR dataset

| Models | | Best results on Pheme1 R/NR (5,791 tweets; 1,969 R, 3,822 NR) | | | |
|---|---|---|---|---|---|
| State-of-the-art models and their split of dataset | | A (%) | P (%) | R (%) | F1 (%) |
| (Wang et al., 2020) | 70:10:20 for train:val:test | 87.56 | 87.62 | 87.65 | 87.64 |
| (Pelrine et al., 2021) | 70:10:20 for train:val:test | - | - | - | 89.4+0.3 |
| Proposed Models [10%:testing, 75% and 25% of rest of the data for training and validation respectively] | | | | | |
| Classifiers | Language models | | | | |
| 4MLP | BERT | 87.304 | 86.416 | 85.448 | 85.929 |
| | RoBERTa | 88.215 | 86.862 | 87.643 | 87.251 |
| | DistilBERT | 87.130 | 86.258 | 85.201 | 85.726 |
| 4L-MLP + Reg+Drop | BERT | 87.040 | 86.017 | 84.684 | 85.345 |
| | RoBERTa | 88.908 | 87.900 | 87.728 | 87.814 |
| | DistilBERT | 87.130 | 86.184 | 85.314 | 85.747 |
| 10L- ResNet-CNN | BERT | 87.840 | 86.612 | 86.093 | 86.352 |
| | RoBERTa | 88.000 | 87.215 | 85.640 | 86.420 |
| | DistilBERT | 87.130 | 86.337 | 85.087 | 85.707 |
| 18-L ResNet-CNN | BERT | 87.040 | 86.017 | 84.684 | 85.345 |
| | RoBERTa | 87.868 | 86.755 | 86.588 | 86.671 |
| | DistilBERT | 87.652 | 86.808 | 85.830 | 86.316 |

Abbreviations: A:Accuracy, P: Precision, R: Recall, F1: F1 score

**Table 4**
Comparison of the proposed method with state-of-the-art techniques for Pheme2 R/NR dataset

| Classifier models | | Best results on Pheme2 R/NR (6425 tweets; 2402 R, 4023 NR) | | | |
|---|---|---|---|---|---|
| State-of-the-art models and their split of dataset | | A (%) | P (%) | R (%) | F1 (%) |
| (Dong et al., 2019b) | Not provided | 83.36 | 83.59 | 99.64 | 90.91 |
| Proposed Models [10%-testing, 75% and 25% of rest of the data for training and validation, respectively] | | | | | |
| 4MLP | BERT | 89.404 | 88.812 | 88.532 | 88.672 |
| | RoBERTa | 90.066 | 89.280 | 89.676 | 89.478 |
| | DistilBERT | 86.342 | 85.553 | 85.983 | 85.767 |
| 4L-MLP + Reg+Drop | BERT | 88.907 | 88.156 | 88.222 | 88.189 |
| | RoBERTa | 90.232 | 89.518 | 89.721 | 89.61 |
| | DistilBERT | 86.424 | 85.819 | 85.006 | 85.411 |
| 10L-ResNet-CNN | BERT | 89.073 | 88.403 | 88.267 | 88.335 |
| | RoBERTa | 89.735 | 88.895 | 89.411 | 89.152 |
| | DistilBERT | 87.252 | 86.478 | 86.282 | 86.380 |
| 18-L ResNet-CNN | BERT | 89.735 | 89.227 | 88.797 | 89.011 |
| | RoBERTa | 90.232 | 89.565 | 89.634 | 89.599 |
| | DistilBERT | 87.086 | 86.279 | 86.150 | 86.214 |

Abbreviations: A:Accuracy, P: Precision, R: Recall, F1: F1 score

Table 5 compares the proposed models' performance on Twitter15 R/NR and Twitter16 R/NR datasets. The results show that all the proposed models perform exceptionally well on Twitter15 R/NR and Twitter16 R/NR datasets. The proposed models using finetuned RoBERTa and BERT LMs outperform the state-of-the-art models ((Ma et al., 2019; Santhoshkumar and Babu, 2020; Wang et al., 2020)) on Twitter15 R/NR dataset. Interestingly on the Twitter16 R/NR dataset, RoBERTa LM which has highest number of trainable parameters obtained the worst accuracy, while DistilBERT LM which has smallest number of trainable parameters achieved the best accuracy. It is expected that size and imbalance of the dataset may have affected the performance of LMs.

For classifying true and false rumours, Tables 6 and 7 presents the performance of the proposed methods on classifying true and false rumours using Pheme2 T/F, Twitter15 T/F, and Twitter16 T/F datasets. It can be observed

**Table 5**
Comparison of the proposed method with state-of-the-art techniques for Twitter15 R/NR and Twitter16 R/NR datasets

| Classifier models | | Best results on Twitter15 R/NR (1490 tweets; 1118 R, 372 NR) | | | | Best results on Twitter16 R/NR (818 tweets; 613 R, 205 NR) | | | |
|---|---|---|---|---|---|---|---|---|---|
| State-of-the-art and split of dataset | | A (%) | P (%) | R (%) | F1 (%) | A (%) | P (%) | R (%) | F1 (%) |
| (Ma et al., 2019) | Not provided | 86.3 | - | - | - | 88.5 | - | - | - |
| (Santhoshkumar and Babu, 2020) | 60:40 for train:test | 86 | - | - | - | 87 | - | - | - |
| Proposed models [10%-testing, 75% and 25% of rest of the data for training and validation, respectively] | | | | | | | | | |
| 4MLP | BERT | 87.586 | 82.711 | 82.711 | 82.711 | *91.358* | 90.211 | **85.861** | **87.982** |
| | RoBERTa | 92.466 | 89.769 | 88.944 | 89.355 | 85.294 | 83.103 | 73.077 | 77.768 |
| | DistilBERT | 85.616 | 81.155 | 76.287 | 78.646 | **91.358** | **92.121** | 84.180 | 87.972 |
| 4L-MLP + Reg+Drop | BERT | 88.966 | 87.266 | 80.551 | 83.774 | 90.123 | 91.205 | 81.680 | 86.18 |
| | RoBERTa | *92.466* | 90.505 | *87.920* | *89.194* | 86.420 | 82.767 | 79.221 | 80.955 |
| | DistilBERT | 85.616 | 80.266 | 78.335 | 79.289 | 91.358 | *92.121* | *84.180* | *87.972* |
| 10L-ResNet-CNN | BERT | 87.586 | 85.882 | 77.610 | 81.537 | 90.123 | 91.205 | 81.680 | 86.18 |
| | RoBERTa | **93.151** | **91.118** | **89.391** | **90.246** | 86.420 | 83.939 | 77.541 | 80.613 |
| | DistilBERT | 86.986 | 82.803 | 79.228 | 80.976 | 91.358 | 92.121 | 84.180 | 87.972 |
| 18-L ResNet-CNN | BERT | 86.538 | 82.181 | 81.624 | 81.902 | 90.123 | 91.205 | 81.680 | 86.18 |
| | RoBERTa | 92.466 | *90.505* | 87.920 | 89.194 | 87.654 | 85.240 | 80.041 | 82.559 |
| | DistilBERT | 88.356 | 83.539 | 84.217 | 83.877 | 91.358 | 92.121 | 84.180 | 87.972 |

Abbreviations: A:Accuracy, P: Precision, R: Recall, F1: F1 score

from the tables that these datasets are relatively small in size but are well balanced. Table 6 presents the experimental results of the proposed models on Pheme2 T/F dataset for identifying true and false rumours in tweets. All the proposed models outperform the state-of-the-art model (Wu et al., 2020a) in all metrics while similar to (Pelrine et al., 2021) in terms of F1 score. This shows that all the proposed models consistently achieve a high performance on Pheme2 T/F dataset. Interestingly, by using 4-layers MLP and ten-layers ResNet-CNN classifier models, BERT and DistilBERT LMs obtained better result as compared to RoBERTa LM which has more trainable parameters. Table 7 compares the proposed models' performance on classifying true and false-rumour on Twitter15 T/F and Twitter16 T/F datasets. It can clearly observed that all the proposed models achieve high performance and outperform the state-of-the-art models ((Lu and Li, 2020; Pelrine et al., 2021)) on both datasets. For Twitter15 T/F dataset, BERT LM with 4-layers MLP classifier performs the best while for Twitter16 T/F dataset, RoBERTa LM with the same classifier obtained the best scores. Therefore, it can be concluded that MLP as a classifier performs better than all other considered classifiers.

The experimental results of the proposed models for classifying true and fake news on Covid T/F as a long-text dataset are presented in Table 8. The results indicate that all the proposed models perform better on Covid T/F dataset as compared to the State-of-the-art methods (Al-Ahmad et al., 2021). The models' performance considerably improves the accuracy by around 5% - 7% and F1-Score by 16%-19% as compared to the state-of-the-art technique. RoBERTa as the largest LM combined with a simple 4-layers MLP classifier performs better than other LMs and classifiers. However, when incorporated with a more complex classifier, such as 18L-Resnet-CNN, RoBERTa's performance slightly decreased. In general, there is no significant difference between the performance of three LMs.

## 4.2. Classification on Combined Datasets

Table 9 and Table 10 present the result on distinguishing rumour/non-rumour and true/false-rumour on combined datasets. As mentioned earlier, this study combined all the same labelled datasets from different sources into a new single dataset to attain a larger and more balanced dataset. It is expected that large data will help to generalise and improve the training of the proposed models. The testing was done with testing set of each individual dataset which was separated before combining the data for training and validation, as well as with combined testing sets taken from all datasets. Therefore, Table 9 compares the models' performance on four different datasets, including Twitter16 R/NR, Twitter15 R/NR, Pheme1 R/NR, Pheme2 R/NR and their combination, and Table 10 compares the models' performance on four different datasets including Twitter16 T/F, Twitter15 T/F, Pheme2 T/F, and their combination.

**Table 6**
Comparison of the proposed method with state-of-the-art techniques for Pheme2 T/F dataset

| Classifier models | | Best results on Pheme2 T/F (1705 tweets; 1067 T, 638 F) | | | |
|---|---|---|---|---|---|
| State-of-the-art techniques and their split of dataset | | A (%) | P (%) | R (%) | F1 (%) |
| (Wu et al., 2020a) | 70:10:20 for train:val:test | 82.46 | 79.08 | 86.24 | 82.5 |
| (Pelrine et al., 2021) | 70:10:20 for train:val:test sets | - | - | - | 93.2 +0.9 |
| Proposed models [10%-testing, 75% and 25% of rest of the data for training and validation, respectively] | | | | | |
| 4MLP | BERT | 90.419 | 90.286 | **89.179** | _89.729_ |
| | RoBERTa | 89.873 | 89.522 | _89.015_ | 89.268 |
| | DistilBERT | 90.323 | **92.113** | 86.866 | _89.413_ |
| 4L-MLP + Reg+Drop | BERT | 89.820 | 89.782 | 88.385 | 89.078 |
| | RoBERTa | 89.241 | 89.251 | 87.891 | 88.566 |
| | DistilBERT | 89.820 | 90.544 | 87.759 | 89.13 |
| 10L-ResNet-CNN | BERT | **91.398** | 91.840 | 89.116 | **90.457** |
| | RoBERTa | 88.623 | 90.162 | 85.859 | 87.958 |
| | DistilBERT | 89.241 | 88.741 | 88.499 | 88.62 |
| 18-L ResNet-CNN | BERT | _90.419_ | _90.614_ | 88.866 | _89.731_ |
| | RoBERTa | 89.241 | 88.741 | 88.499 | 88.62 |
| | DistilBERT | 87.425 | 87.140 | 85.836 | 86.483 |

Abbreviations: A:Accuracy, P: Precision, R: Recall, F1: F1 score

It can be observed from the results presented in Table 9 and Table 10 that the proposed models perform well and obtain high performance as compared to the results obtained on individual datasets presented in Tables 3-8. This is due to the fact that more training and validation data is made available for LMs and classifiers. There was no significant difference in performance for light or large LMs with a simple or more complex classifier. The difference between all the models on combined dataset was around 1% to 2%. This shows that combining data can increase generalisation during the training process and help stabilise the proposed model's performance.

**Table 7**
Comparison of the proposed method with state-of-the-art techniques for Twitter15 T/F and Twitter16 T/F datasets

| Classifier models | | Best results on Twitter15 T/F (754 tweets; 374 T, 370 F) | | | | Best results on Twitter16 T/F (410 tweets; 205 T, 205 F) | | | |
|---|---|---|---|---|---|---|---|---|---|
| State-of-the-art techniques and their split of dataset | | A (%) | P (%) | R (%) | F1 (%) | A (%) | P (%) | R (%) | F1 (%) |
| (Lu and Li, 2020) | 70:30 for train:test | 87.67 | 82.57 | 82.95 | 82.50 | 90.84 | 75.94 | 76.32 | 75.93 |
| (Pelrine et al., 2021) | 70:10:20 for train:val:test | - | - | - | 94.4$\pm$0.8 | - | - | - | 95.7$\pm$2.8 |
| Proposed models [10%-testing, 75% and 25% of rest of the data for training and validation, respectively] | | | | | | | | | |
| 4MLP | BERT | **97.015** | **97.059** | **97.143** | **97.101** | 92.308 | 93.182 | 92.500 | 92.840 |
| | RoBERTa | _97.015_ | _97.009_ | _97.009_ | _97.009_ | **97.436** | **97.500** | **97.500** | **97.500** |
| | DistilBERT | 95.522 | 95.499 | 95.580 | 95.539 | 88.889 | 88.889 | 88.259 | 88.573 |
| 4L-MLP + Reg+Drop | BERT | 95.000 | 95.455 | 95.000 | 95.227 | 92.308 | 93.182 | 92.500 | 92.840 |
| | RoBERTa | _97.015_ | 97.009 | 97.009 | 97.009 | _97.436_ | _97.500_ | _97.500_ | _97.500_ |
| | DistilBERT | 94.030 | 94.018 | 94.018 | 94.018 | 88.889 | 88.889 | 88.259 | 88.573 |
| 10L-ResNet-CNN | BERT | 95.522 | 95.714 | 95.714 | 95.714 | 94.872 | 95.455 | 94.737 | 95.095 |
| | RoBERTa | 97.015 | 97.009 | 97.009 | 97.009 | 97.436 | _97.500_ | _97.500_ | _97.500_ |
| | DistilBERT | 94.030 | 94.018 | 94.018 | 94.018 | _97.436_ | 97.619 | 97.368 | 97.493 |
| 18-L ResNet-CNN | BERT | 92.537 | 93.243 | 92.857 | 93.05 | 94.872 | 94.868 | 94.868 | 94.868 |
| | RoBERTa | 95.522 | 95.499 | 95.580 | 95.539 | 97.436 | 97.500 | 97.500 | 97.500 |
| | DistilBERT | 97.015 | _97.059_ | _97.143_ | _97.101_ | 88.889 | 88.889 | 88.259 | 88.573 |

Abbreviations: A:Accuracy, P: Precision, R: Recall, F1: F1 score

**Table 8**
Comparison of the proposed method with state-of-the-art techniques for Covid T/F dataset

| Classifier models | | Best results on Covid T/F (3119 News; 2061 T, 1058 F) | | | |
|---|---|---|---|---|---|
| State-of-the-art techniques ((Al-Ahmad et al., 2021)) Splitting data strategies are not provided | | A (%) | P (%) | R (%) | F1 (%) |
| k-NN-BSSA | | 72.610 | 59.620 | 59.740 | 59.68 |
| k-NN-BPSO | | 73.120 | 61.980 | 53.780 | 57.59 |
| k-NN-BGA | | 75.430 | 66.220 | 56.330 | 60.88 |
| k-NN | | 70.650 | 56.420 | 59.360 | 57.85 |
| J48 | | 72.290 | 59.600 | 56.900 | 0.58.22 |
| RF | | 70.410 | 63.420 | 30.150 | 40.87 |
| SVM | | 70.750 | 56.650 | 58.790 | 57.7 |
| Proposed models [10%-testing, 75% and 25% of rest of the data for training and validation, respectively] | | | | | |
| 4MLP | BERT | 80.192 | 78.640 | 77.024 | 77.824 |
| | RoBERTa | _82.026_ | _79.961_ | _79.518_ | _79.739_ |
| | DistilBERT | 80.511 | 79.612 | 76.438 | 77.993 |
| 4L-MLP + Reg+Drop | BERT | 81.046 | 79.223 | 77.345 | 78.273 |
| | RoBERTa | **82.353** | **80.549** | 79.286 | **79.913** |
| | DistilBERT | 80.511 | _79.788_ | 76.229 | 77.968 |
| 10L-ResNet-CNN | BERT | 81.046 | 79.223 | 77.345 | 78.273 |
| | RoBERTa | 81.046 | 78.730 | 79.258 | 78.993 |
| | DistilBERT | 80.511 | _79.982_ | 76.021 | 77.951 |
| 18-L ResNet-CNN | BERT | _81.699_ | 79.474 | **79.750** | _79.612_ |
| | RoBERTa | 81.046 | 78.722 | _79.497_ | 79.108 |
| | DistilBERT | 80.192 | 79.679 | 75.567 | 77.569 |

Abbreviations: A:Accuracy, P: Precision, R: Recall, F1: F1 score

**Table 9**

Comparison of the proposed method with State-of-the-art techniques for Twitter16 R/NR, Twitter15 R/NR, Pheme1 R/NR, Pheme2 R/NR and Combined R/NR datasets

| Classifier models | | Twitter16 R/NR (613/205) | | Twitter15 R/NR (1,118/372) | | Pheme1 R/NR (1,969/3,822) | | Pheme2 R/NR (2,402/4,023) | | Combined R/NR (4,133/4,600) | |
|---|---|---|---|---|---|---|---|---|---|---|---|
| | | A% | F1 (%) | A (%) | (F1%) | A (%) | F1 (%) | A (%) | F1 (%) | A (%) | F1 (%) |
| 4MLP | BERT | 91.358 | **87.982** | 87.586 | 82.711 | 88.215 | 87.251 | *90.066* | *89.478* | 88.449 | 88.479 |
| | RoBERTa | 85.294 | 77.768 | 92.466 | 89.355 | 87.304 | 85.929 | 89.404 | 88.672 | **90.099** | 90.092 |
| | DistilBERT | **91.358** | 87.972 | 85.616 | 78.646 | 87.130 | 85.726 | 86.342 | 85.726 | 88.999 | 88.991 |
| 4L-MLP+ Reg+Drop | BERT | 90.123 | 86.18 | 88.966 | 83.774 | **88.908** | **87.814** | 90.232 | 89.613 | 88.779 | 88.831 |
| | RoBERTa | 86.420 | 80.955 | *92.466* | *89.194* | 87.040 | 85.345 | 88.907 | 88.189 | 90.099 | *90.091* |
| | DistilBERT | *91.358* | *87.972* | 85.616 | 79.289 | 87.130 | 85.747 | 86.424 | 85.411 | 89.109 | 89.101 |
| 10L-ResNet-CNN | BERT | 90.123 | 86.180 | 87.586 | 81.537 | *88.000* | 86.420 | 89.735 | 89.152 | 88.889 | 88.945 |
| | RoBERTa | 86.420 | 80.613 | **93.151** | **90.246** | 87.840 | 86.352 | 89.073 | 88.335 | *90.099* | **90.107** |
| | DistilBERT | 91.358 | 87.972 | 86.986 | 80.976 | 87.130 | 85.707 | 87.252 | 86.380 | 89.219 | 89.210 |
| 18-L ResNet-CNN | BERT | 90.123 | 86.180 | 86.538 | 81.902 | 87.868 | *86.671* | 90.232 | 89.599 | 88.559 | 88.593 |
| | RoBERTa | 87.654 | 82.559 | 92.466 | 89.194 | 87.040 | 85.345 | 89.735 | 89.011 | 90.096 | 90.090 |
| | DistilBERT | 91.358 | 87.972 | 88.356 | 83.877 | 87.652 | 86.316 | 87.086 | 86.214 | 88.991 | 88.991 |

Abbreviations: A:Accuracy, F1: F1 score

**Table 10**

Comparison of the proposed method with state-of-the-art techniques for Twitter16 T/F, Twitter15 T/F, Pheme T/F and Combined T/F datasets

| Classifier models | | Twitter16 T/F (410 tweets) | | Twitter15 15 T/F (754 tweets) | | Pheme2 T/F (1705 tweets) | | Combined T/F (2869 tweets) | |
|---|---|---|---|---|---|---|---|---|---|
| | | A% | F1 (%) | A (%) | F1 (%) | A (%) | F1 (%) | A (%) | F1 (%) |
| 4MLP | BERT | 92.308 | 92.840 | **97.015** | **97.101** | 90.419 | *89.729* | **93.624** | **93.567** |
| | RoBERTa | **97.436** | **97.500** | 97.015 | 97.009 | 89.873 | 89.268 | 92.617 | 92.459 |
| | DistilBERT | 88.889 | 88.573 | 95.522 | 95.539 | *90.323* | 89.413 | 92.883 | 92.814 |
| 4L-MLP+ Reg+Drop | BERT | 92.308 | 92.840 | 95.000 | 95.227 | 89.820 | 89.078 | 93.624 | 93.567 |
| | RoBERTa | 97.436 | 97.500 | *97.015* | *97.009* | 89.241 | 88.566 | 92.282 | 92.124 |
| | DistilBERT | 88.889 | 88.573 | 94.030 | 94.018 | 89.820 | 89.130 | 92.953 | 92.778 |
| 10L-ResNet-CNN | BERT | 94.872 | 95.095 | 95.522 | 95.714 | **91.398** | **90.457** | 92.617 | 92.421 |
| | RoBERTa | 97.436 | 97.500 | 97.015 | 97.009 | 88.623 | 87.958 | 92.100 | |
| | DistilBERT | *97.436* | 97.493 | 94.030 | 94.018 | 89.241 | 88.620 | *92.953* | 92.778 |
| 18-L ResNet-CNN | BERT | 94.872 | 94.868 | 92.537 | 93.050 | 90.419 | 89.731 | 92.953 | *92.930* |
| | RoBERTa | 97.436 | 97.500 | 95.522 | 95.539 | 89.241 | 88.620 | 92.617 | 92.459 |
| | DistilBERT | 88.889 | 88.573 | 97.015 | 97.101 | 87.425 | 86.483 | 92.282 | 92.100 |

Abbreviations: A:Accuracy, F1: F1 score

# 5. Conclusion

This study demonstrates that incorporating a BERT-based finetuned language model as an encoder and a neural network as a classifier can improve the accuracy of misinformation detection. The proposed models consistently achieved high-performance on both short and long-text datasets and outperformed the current state-of-the-art models.

The experimental results show that the performance of proposed models improved significantly when trained and tested on larger amount of data which was achieved by combining all the considered datasets. It was also observed that there was no significant difference between different LMs, whether large LM or a light LM. The performance differ by 1-2% only between different LMs. Hence, it can be concluded that combining datasets helps to evaluate the generalised performance of the proposed model.

The results also indicate that a sophisticated model does not guarantee a better outcome. Many studies implied that using models with large number of trainable parameters improve performance. However, a simple 4-layers MLP classifier with simple LM can perform better than combination of complex classifiers (ResNet-CNN) and LMs (BERT or DistillBERT). The difference of results obtained from RoBERTa, BERT and DistilBERT with different classifiers were insignificant: only 1% to 3% difference. Therefore, it is suggested that in real-world scenarios training time, cost and computational complexity of the models should be considered carefully rather than small improvements in resuls. Future studies should examine the interaction between datasets, algorithm models, and splitting of data to understand the solution in more comprehensive manner.

---

## Acknowledgment

The authors express gratitude to the Sebelas Maret University for exceptional support as a sponsor of this research.

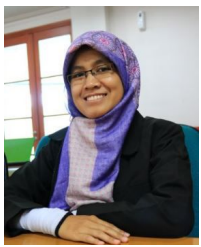

Rini Anggrainingsih received her bachelor degree and masters from Diponegoro University and Gadjahmada University, respectively. She is currently working as academic staff at Sebelas Maret University while pursuing a PhD at The University of Western Australia. Her current research interest are false information detection on the online platforms, and other natural language processing tasks to improve info-surveillance.

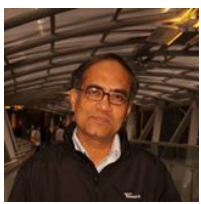

Amitava Datta (A'04–M'10) received the M.Tech. and Ph.D. degrees from IIT Madras in 1988 and 1992, respectively. He did the post-doctoral research at the Max Planck Institut für Informatik, Germany, and the University of Freiburg, Germany. He joined the University of New England in 1995 and The University of Western Australia in 1998, where he is currently a Professor with the School of Computer Science and Software Engineering. His current research interests are in optical computing, data mining, bioinformatics, and social network analysis. He has authored over 150 papers in various international journals and conference proceedings, including the IEEE Transactions on Computers, the IEEE Transactions on Parallel and Distributed Systems, the IEEE Transactions on Visualization and Computer Graphics, the IEEE Transactions on Mobile Computing, the IEEE/ACM Transactions on Networking, the IEEE Transactions on Computational Social Systems and the IEEE Transactions on Systems, Man, and Cybernetics.

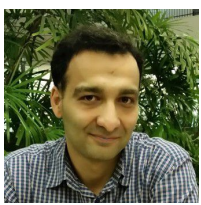

Dr. Ghulam Mubashar Hassan received his B.Sc. Electrical and Electronics Engineering degree (with honors) from University of Engineering and Technology (UET) Peshawar, Pakistan. He completed his MS in Electrical and Computer Engineering from Oklahoma State University USA and his PhD from The University of Western Australia (UWA) in Joint Schools of Computer Science & Software Engineering and Civil & Resource Engineering. He was valedictorian and received many awards for his PhD. Currently, he is working in UWA and previously he worked in UET Peshawar and King Saud University. His research interests are multidisciplinary problems which include using artificial intelligence, machine learning, pattern recognition, optimization in different fields of engineering and education.